%% file: 0-Main.tex
\documentclass[letterpaper]{article} % DO NOT CHANGE THIS
\usepackage{aaai25}  % DO NOT CHANGE THIS
\usepackage{times}  % DO NOT CHANGE THIS
\usepackage{helvet}  % DO NOT CHANGE THIS
\usepackage{courier}  % DO NOT CHANGE THIS
\usepackage[hyphens]{url}  % DO NOT CHANGE THIS
\usepackage{graphicx} % DO NOT CHANGE THIS
\usepackage{natbib}  % DO NOT CHANGE THIS AND DO NOT ADD ANY OPTIONS TO IT
\usepackage{caption} % DO NOT CHANGE THIS AND DO NOT ADD ANY OPTIONS TO IT 
\usepackage{algorithm}
\usepackage{algorithmic}

\usepackage{color}

\usepackage{bm}
\usepackage{enumitem}

\usepackage{amsmath,amsfonts}

\usepackage{subfig}
\usepackage{float}

\usepackage{makecell}

\usepackage{xcolor}
\usepackage{colortbl}
\usepackage{array}
\usepackage{multirow}
\usepackage{booktabs}
\usepackage{threeparttable}

\newcommand{\rc}[1]{\textcolor{red}{#1}}
\newcommand{\M}{RAGPT}

\newlength\savedwidth

\newlength\savewidth

\usepackage{newfloat}
\usepackage{listings}
\DeclareCaptionStyle{ruled}{labelfont=normalfont,labelsep=colon,strut=off} % DO NOT CHANGE THIS
\floatstyle{ruled}
\newfloat{listing}{tb}{lst}{}
\floatname{listing}{Listing}
\title{Retrieval-Augmented Dynamic Prompt Tuning for  Incomplete \\ Multimodal Learning}

\author {
    Jian Lang\textsuperscript{\rm 1}\equalcontrib,
    Zhangtao Cheng\textsuperscript{\rm 1}\equalcontrib,
    Ting Zhong\textsuperscript{\rm 1,2},
    Fan Zhou\textsuperscript{\rm 1,2}\thanks{Corresponding author.}
}
\affiliations {
    \textsuperscript{\rm 1}University of Electronic Science and Technology of China, Chengdu, Sichuan, China\\
    \textsuperscript{\rm 2}Kash Institute of Electronics and Information Industry, Kashgar, Xinjiang, China\\
    jian\_lang@std.uestc.edu.cn, zhangtao.cheng@outlook.com, zhongting@uestc.edu.cn, fan.zhou@uestc.edu.cn
}

\usepackage{bibentry}
\begin{document}

\maketitle

\begin{abstract}

Multimodal learning with incomplete modality is practical and challenging. Recently, researchers have focused on enhancing the robustness of pre-trained MultiModal Transformers (MMTs) under missing modality conditions by applying learnable prompts. However, these prompt-based methods face several limitations: (1) incomplete modalities provide restricted modal cues for task-specific inference, (2) dummy imputation for missing content causes information loss and introduces noise, and (3) static prompts are instance-agnostic, offering limited knowledge for instances with various missing conditions. To address these issues, we propose \M, a novel \textbf{R}etrieval-\textbf{A}u\textbf{G}mented dynamic \textbf{P}rompt \textbf{T}uning framework. \M~comprises three modules: (I) the multi-channel retriever, which identifies similar instances through a within-modality retrieval strategy, (II) the missing modality generator, which recovers missing information using retrieved contexts, and (III) the context-aware prompter, which captures contextual knowledge from relevant instances and generates dynamic prompts to largely enhance the MMT’s robustness. 
Extensive experiments conducted on three real-world datasets show that \M~consistently outperforms all competitive baselines in handling incomplete modality problems. 
\end{abstract}

\section{Introduction}
\label{Introduction}
\input{1-Introduction.tex}

\section{Related Work}
\label{Related Work}
\input{2-Related_Work.tex}

\section{Methodology}
\label{Methodology}
\input{4-Methodology.tex}

\section{Experiments}
\label{Experiments}
\input{5-Experiment.tex}

\section{Conclusion}
\label{Conclusion}
\input{6-Conclusion.tex}

\section{Acknowledgments}
This work was supported by National Natural Science Foundation of China (Grant No.62176043, No.62072077, and No.U22A2097), and Kashgar Science and Technology Bureau (Grant No.KS2023025).

\bibliography{aaai25}

\end{document}

%% file: 1-Introduction.tex
Multimodal learning has emerged as a critical paradigm in both research and industry, demonstrating broad application potential in areas such as healthcare assistance \cite{ghosh2024clipsyntel} and malicious content detection \cite{kiela2020hateful}. However, most successful methods typically assume that the completeness of all modalities is essential during both training and inference phases. In reality, factors such as malfunctioning sensors and privacy concerns often make it infeasible to collect complete modalities \cite{ma2021smil}. As a result, the challenge of incomplete modalities significantly impacts the reliability, accuracy, and safety of multimodal models in practical applications \cite{woo2023towards, cheng2024retrieval}.

To address this challenge, researchers have developed various robust multimodal methods that are broadly categorized into three groups: \textit{(1) Joint learning methods} \cite{wang2023multi, yao2024drfuse}, \textit{(2) Cross-modal generation methods} \cite{ma2021smil, woo2023towards},
and \textit{(3) Prompt-based methods} \cite{lee2023multimodal, jang2024towards}. For joint learning methods, they heavily rely on the selection of similarity measures and require filling missing-modality inputs with masking values, resulting in the loss of critical information and the introduction of noise into the models \cite{wang2024gradient}. 
Cross-modal generation methods inevitably face modality heterogeneity issues and incur limited reconstruction quality. 
Recently, prompt-based methods have gained significant attention due to the rise of powerful pre-trained MultiModal Transformers (MMTs).
These methods leverage prompt-tuning techniques to effectively transfer the capabilities of MMTs pre-trained on complete multimodal datasets to tasks involving missing modalities, achieving remarkable performance and making them a dominant trend in incomplete multimodal learning. 

\begin{figure}[t]
    \centering    
    {\includegraphics[width=1\columnwidth]{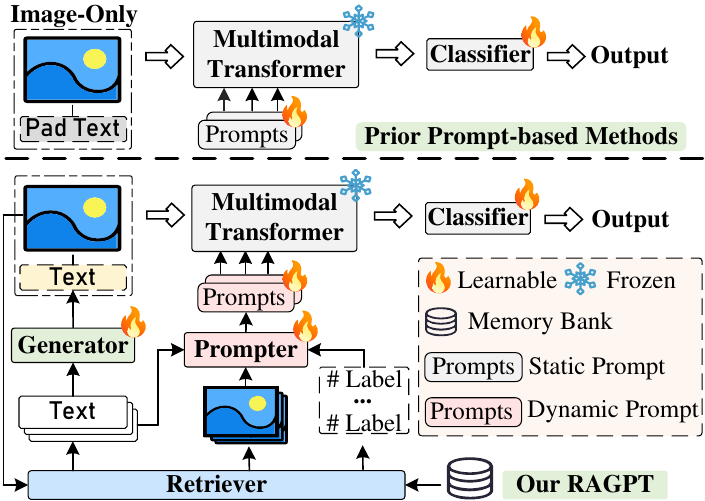}}  
    \caption{Prior prompt-based methods \textit{vs.} our \M~in tackling incomplete multimodal learning. }
    \label{fig:description}
\end{figure}
 
However, for incomplete modalities, prompt-based methods typically use the available modalities as the only cue to fulfill task-specific objectives through prompt learning (see Fig.~\ref{fig:description}). Despite their progress, these methods often struggle in severe missing-modality scenarios due to several unresolved issues inherent in their design:
\textbf{(1)} Remaining modalities typically provide restricted modal information, which fails to effectively address specific tasks when the missing modality contains crucial modal cues. \textbf{(2)} Modal incomplete inputs are often filled with dummy values (e.g., empty strings/pixels for texts/images), which may introduce noise, leading to degraded performance \cite{ma2022multimodal}. \textbf{(3)} The prompt tokens are shared across any inputs and therefore are instance-agnostic. Thus, this static prompt-tuning is not well-suited for real multimodal instances, as instances with different types of missing modalities belong to distinct distributions. Additionally, static prompts typically provide limited knowledge for both missing- and full-modality instances.  
Therefore, these observations motivate us to design a universal prompt-tuning strategy to enhance the pre-trained MMT's robustness for incomplete modalities.

To address these issues, we draw inspiration from the human ability to learn through observation, which involves mastering skills by observing relevant subjects rather than attempting to memorize every subject \cite{hodges2007modelled}. As shown in Fig.~\ref{fig:description}, we leverage this cognitive principle to address the challenge of missing modalities. Our core idea is to retrieve relevant multimodal contents and utilize them as prompts to enhance the robustness of pre-trained MMT in both missing- and full-modality scenarios. Intuitively, for instances with missing modalities, appending multimodal content from similar instances can provide contextual knowledge relevant to the missing modality and improve task-specific predictions.

To this end, we propose \textbf{\M}, a novel \textbf{R}etrieval-\textbf{A}u\textbf{G}mented dynamic \textbf{P}rompt \textbf{T}uning framework to adaptively enhance the robustness of pre-trained MMT in both missing- and full-modality scenarios. 
Fundamentally, we reformulate incomplete modality learning in a principled retrieve-and-prompt manner and maintain a model-agnostic design that facilitates seamless integration with various prompt-based models. \M~includes three modules: multi-channel retriever, missing modality generator, and context-aware prompter. 
During retrieval, we propose a universal multi-channel retrieval strategy that disentangles multimodal representations into unimodal components, facilitating the retrieval of similar samples based on within-modality similarities for missing- and full-modality scenarios. 

Next, the missing modality generator comprises a learnable filter to approximate the missing information.
Beyond traditional reconstruction techniques, which suffer from modality gaps during the cross-modal generation, this generator realizes \textit{intra-modal reconstruction} by leveraging information from the retrieved samples that belong to the same modality as the missing one to recover the missing content. 
Moreover, this design enriches the missing-modality representation, ensuring alignment with the complete-modality input format of pre-trained MMTs during the pre-training phase.
Finally, the context-aware prompter identifies the semantic correlations between the target and retrieved instances, producing dynamic multimodal prompts tailored to different inputs. These prompts facilitate the adaptive refinement of modality features in both missing- and full-modality scenarios, thereby enhancing the robustness of the pre-trained models. 
We insert these modules into the pre-trained MMTs to achieve a more accurate representation for both missing- and full-modality data. 
Following are our main contributions:

\noindent $\bullet$ To our best knowledge, this is the first retrieval-augmented paradigm for incomplete modalities. We reveal that prior prompt-based methods suffer from issues related to dummy padding and static prompts, which drastically degrade performance in severe missing-modality cases.

\noindent $\bullet$ To address these issues, we propose \M, pioneering a retrieval-augmented dynamic prompt-tuning framework that bridges target and relevant instances, recovers missing modality, and generates dynamic prompts to enhance the MMT's robustness in diverse missing-modality situations.  

\noindent $\bullet$ We conduct extensive experiments on three real-world datasets to evaluate \M~in comparison with 9 competitive baselines and the results confirm \M's effectiveness in addressing missing-modality issues. The code of our work and prompt-based baselines is available at \text{https://github.com/Jian-Lang/RAGPT}.

%% file: 2-Related_Work.tex
\noindent \textbf{Incomplete Multimodal Learning}. Researchers have developed various methods for incomplete multimodal learning, which can be divided into three groups: \textit{(1) Joint learning methods} \cite{zhao2021missing, wang2023multi, yao2024drfuse} focus on distilling complex correlations from complete modalities to tackle missing-modality data. However, these methods require filling modality-incomplete inputs with masking values, which may cause unexpected behavior and introduce additional noise. \textit{(2) Cross-modal generation methods} \cite{lee2019audio, yuan2021transformer} primarily reconstruct the missing content by using remaining modalities. Researchers \cite{ma2021smil, woo2023towards} directly employ VAE to generate the missing-modality based only on available modalities. Consequently, these methods inevitably face modality heterogeneity problems. \textit{(3) Prompt-based methods} \cite{lee2023multimodal, jang2024towards} introduces learnable prompts to help pre-trained MMTs address incomplete modalities.

However, prompt-based methods are constrained by the dummy imputation and static prompting strategy, resulting in performance bottlenecks. In contrast, our \M~captures contextual knowledge from retrieved instances to recover the missing content and generate dynamic prompts to enhance the MMT's robustness for missing modalities.

\begin{figure*}[t]
  \centering
  \includegraphics[width=1\textwidth]{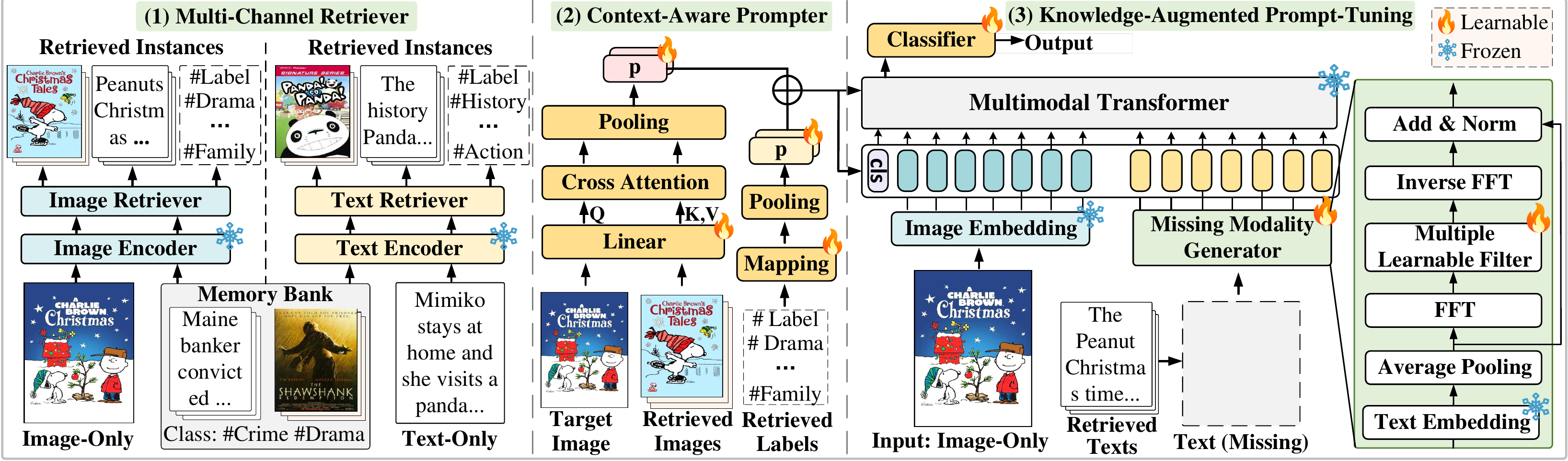}
  \caption{Overall framework of \M. (1) The multi-channel retriever identifies similar instances through a within-modality retrieval strategy; (2) The context-aware prompter captures contextual knowledge from relevant instances and generates dynamic prompts; (3) The knowledge-augmented prompt-tuning process first recovers the missing content by using a missing-modality generator and then performs dynamic prompt-tuning on the pre-trained MMT for final prediction.}
  \label{fig:model}
\end{figure*} 

\noindent \textbf{Prompt Learning}. Prompt learning \cite{liu2023pre} utilizes 
a small number of learnable prompt parameters added to the input of pre-trained transformers, facilitating adjustments to the pre-trained models for alignment with downstream tasks. It has been successfully applied to various domains, such as visual identity~\cite{khattak2023maple, lee2023multimodal} and social network analysis~\cite{zhou2021survey, xu2021casflow, zhong2024predicting, cheng2024information, cheng2023enhancing}. 
Following the success of prompt learning in NLP tasks \cite{Xiang2021Prefix}, recent works have attempted to explore its application in multimodal learning \cite{zhou2022learning}. 
For instance, MaPLe \cite{khattak2023maple} introduces a soft prompt appended to the hidden representations of MMTs, resulting in significant improvements in few-shot image recognition. 
For incomplete multimodal learning, MAPs \cite{lee2023multimodal} and MSPs \cite{jang2024towards} design various prompts to fine-tune pre-trained MMTs, enabling them to adapt effectively to missing-modality scenarios. However, these prompts are instance-agnostic and provide limited information for both missing- and full-modality data. In contrast, the context-aware prompter in \M~captures rich contextual knowledge from relevant instances, alleviating the drawbacks associated with instance-agnostic prompts.

%% file: 4-Methodology.tex
\noindent \textbf{Problem Definition}: In this paper, we consider a multimodal dataset incorporating two modalities. Formally, we define $D = \{D^f,D^{m}\}$ to represent the multimodal dataset. Here, $D^f = \{(\bm{x}_i^{1}, \bm{x}_i^{2}, y_i)\}_{i=1}^{N^{f}}$ represents the modality-complete subset, where $y_i$ is the class label of $i$-th instance. $\bm{x}_i^{1}$ and $\bm{x}_i^{2}$ denote two modalities (e.g., texts and images). $N^{f}$ is the total number of instances in the subset $D^{f}$. Conversely, $D^{m} = \{(\bm{x}_i^{1}, \_\_, y_i) \lor (\_\_, \bm{x}_i^{2}, y_i)\}_{i=1}^{N^{m}}$ is a modality-incomplete subset, where ``$\_\_$'' indicates a missing modality and $N^{m}$ is the number of missing-modality data in $D^{m}$. The objective of the task is to enhance model robustness in cases where modalities are missing during both training and testing phases.

Fig.~\ref{fig:model} presents the key components and their relationships in \M. The following sections delve into the specifics of each component and their respective implementations.

\subsection{Multi-Channel Retriever}

In this section, we design a unified multi-channel retriever to identify similar modal content for queries within their respective modalities by using within-modality similarities. 

\noindent \textbf{Memory Construction} To store high-quality semantic information as prior knowledge, we define the memory $\mathcal{B}$, which encodes multimodal instances using a collection of $(\text{image}, \text{text}, \text{label})$ triples. 

\noindent \textbf{Multi-Channel Retrieval} 
To adapt diverse missing- and full-modality scenarios, we develop a Multi-Channel Retriever (MCR) that effectively retrieves relevant instances through a unified retrieval architecture.  
Specifically, for the text-missing channel, the MCR employs the image representation as a query to identify top-$K$ similar images and incorporates the associated texts to create multimodal instances. 
For complete modalities, the MCR utilizes both the image and text to search relevant texts and images, respectively, thereby creating multimodal instances. 

Specifically, in the text-level branch, the MCR first tokenizes the text $\bm{x}^{1}_i$ in the target instance $\mathcal{T}_i$ into $n$ word tokens and then projects them into word embedding $\mathcal{W}_i \in \mathbb{R}^{n \times d_t}$, where $d_t$ is the dimension of word embedding. Next, the embedding $\mathcal{W}_i$ is fed into a pre-trained textual encoder (e.g., CLIP textual encoder \cite{radford2021learning}) $\Psi_{t}(\cdot)$ to obtain text representation, represented as $\mathbf{E}^{t}_i =  \Psi_{t}(\mathcal{W}_i) \in \mathbb{R}^{d_t}$. Subsequently, the MCR utilizes the text query $\mathbf{E}^{t}_i$ to calculate similarity scores with the text representation $\mathbf{E}^{t}_{r}$ from the memory $\mathcal{B}$, enabling the identification of the top-$K$ textually similar instances $\mathcal{C}^{\mathcal{R}}_i$:
\begin{equation}\label{Eq:retrieval}
\mathcal{C}^{\mathcal{R}}_i = \underset{r \in \mathcal{B}}{\mathrm{Top\text{-}}K} (\frac{{\mathbf{E}^{t}_i}^{\top} \mathbf{E}_r^{t}}{\|\mathbf{E}_i^{t}\| \cdot \|\mathbf{E}_r^{t}\|}).
\end{equation}

For the vision content, the MCR first divides the image $\bm{x}_i^{2}$ into $m$ non-overlapping patches and then projects them into a sequence of patch tokens $\mathcal{V}_{i} \in \mathbb{R}^{m\times d_v}$. Next, these tokens $\mathcal{V}_{i}$ are input into a pre-trained vision encoder (e.g., CLIP vision encoder \cite{radford2021learning}) $\Psi_{v}(\cdot)$ to obtain vision query $\mathbf{E}^{v}_{i} \in \mathbb{R}^{d_v}$. Finally, the retrieval process for searching top-$K$ vision content is the same as defined in Eq.~\ref{Eq:retrieval}. After retrieval, the top-$K$ instances $\mathcal{C}^{\mathcal{R}}_i = \{c^{r_1}_i,\cdots, c^{r_K}_i\}$ can be readily obtained. Each retrieved instance $c^{r_k}_i$ contains the $(\text{image}, \text{text}, \text{label})$ triplet. 
The retrieved top-$K$ instances provide auxiliary context, guiding the recovery of missing content in the target instance and improving task-specific predictions.

\subsection{Context-Aware Prompter}
To explicitly capture expressive contextual information and enhance robustness of pre-trained MMTs against missing-modality issues, we design a Context-Aware Prompter (CAP) that constructs text-, vision-, and label-level dynamic prompts from the retrieved instances $\mathcal{C}^{\mathcal{R}}_{i}$. 
For text-level prompts, the CAP fuses the reference textual features in $\mathcal{C}_{i}^{\mathcal{R}}$ and aligns textual embedding in $\mathcal{T}_i$ through a simple network. 
Specifically, the CAP first tokenizes and projects the texts $\bm{x}^{1}_{i}$ and $\{\bm{x}^{1, r_k}_{i}\}_{k=1}^K$ into word embeddings $\mathcal{W}_{i} \in \mathbb{R}^{n \times d_t}$ and $\mathcal{W}^{\mathcal{R}}_i 
 = \{\mathcal{W}_{i}^{r_k} \}_{k=1}^K \in \mathbb{R}^{K \times n \times d_t}$. 
Subsequently, the word embedding $\mathcal{W}_{i}$ is used as the query to interact with the retrieved text features $\{\mathcal{W}_{i}^{r_k} \}_{k=1}^K$ via a cross-attention block to facilitate comprehension of context, thereby generating the text-level comprehensive representation $\tilde{\mathbf{P}}^{t}_i \in \mathbb{R}^{n \times d_t}$: 
\begin{gather}\label{Eq:cross-attention}
\tilde{\mathbf{P}}^{t}_i = \text{Att}\left(f^Q_t(\mathcal{W}_{i}), f^K_t(\mathcal{W}^{\mathcal{R}}_i), f^V_t(\mathcal{W}^{\mathcal{R}}_i)\right), \\
\text{Att}(\mathbf{Q}, \mathbf{K}, \mathbf{V}) = \operatorname{Softmax}\left(\frac{\mathbf{Q} \mathbf{K}^T}{\sqrt{d}}\right) \mathbf{V},
\end{gather}
where $f^Q_t(.), f^K_t(.), f^V_t(.)$ denote the query, key, and value projection functions, respectively.
For vision-level prompts, the CAP uses the same process to interact the vision patch tokens $\mathcal{V}_{i} \in \mathbb{R}^{m \times d_v}$ with the retrieved patch tokens $\mathcal{V}^{\mathcal{R}}_i \in \mathbb{R}^{K \times m \times d_v}$ 
to obtain the vision-level representation $\tilde{\mathbf{P}}^{v}_i \in \mathbb{R}^{m \times d_v}$. Then, 
the CAP employs an adaptive pooling strategy to obtain the final context-aware prompts $\mathbf{P}^{t}_i \in \mathbb{R}^{l \times d_t}$ and $\mathbf{P}^{v}_i \in \mathbb{R}^{l \times d_v}$, where $l$ is the prompt length. For label-level prompts, the CAP yields a label embedding matrix $\tilde{\mathbf{P}}_{i}^{l} \in \mathbb{R}^{C \times d}$ to encode $C$ class labels, where $d$ is an adjustable dimension. Given retrieved labels, the CAP performs a look-up operation on embedding matrix $\tilde{\mathbf{P}}_{i}^{l}$ and obtains each label embedding. Next, the CAP averages $K$ label embeddings and generates label-level prompts $\mathbf{P}_{i}^{l} \in \mathbb{R}^{d}$.

\subsection{Knowledge-Augmented Prompt-Tuning}
In this process, we first utilize the retrieved modal information to approximate the missing content through a missing modality generator. Next, we perform dynamic prompt-tuning on the pre-trained MMT (e.g., ViLT \cite{kim2021vilt}) to enhance task-specific inference.

\noindent \textbf{Missing Modality Generator} Existing reconstruction methods \cite{ma2021smil} address missing-modality issues by recovering missing content through available modalities. However, these methods often overlook the modal heterogeneity issue and rely on complex generative structures. Based on these observations, we propose a Missing Modality Generator (MMG) that recovers the missing modality through an ``intra-modal reconstruction". The MMG leverages retrieved content of the same modality as the missing one and incorporates a learnable filter layer to effectively approximate the missing modality in a simpler but effective manner. Specifically, given the text-missing instance $\mathcal{T}_i$, the MMG employs a non-parametric strategy to average all text embeddings $\mathcal{W}^{\mathcal{R}}_i = \{\mathcal{W}_{i}^{r_k} \}_{k=1}^K$ from retrieved instances $\mathcal{C}^{\mathcal{R}}_i$, thereby obtaining textual representation $\bar{\mathcal{W}_i} \in \mathbb{R}^{n \times d_t}$ to approximate the missing modality.

Considering potential noise in comprehensive textual representation $\bar{\mathcal{W}_i}$, the MMG introduces a simple learnable filter block (i.e., MLP-based filter \cite{zhou2022filter}) to efficiently refine textual features $\bar{\mathcal{W}_i}$ by removing noise. Specifically, the MMG employs the Fast Fourier Transform (FFT) along the textual dimension. This operation transforms the text context representation $\bar{\mathcal{W}_i}$ into the frequency domain:

\begin{equation}
 \mathbf{Z}_i = \mathcal{F}(\bar{\mathcal{W}_i}) \in \mathbb{C}^{n \times d_t}, 
\end{equation}
where $\mathcal{F}(\cdot)$ denotes the one-dimensional FFT, and $\mathbf{Z}_i$ is the spectrum of $\bar{\mathcal{W}_i}$. The MMG then modulates the spectrum by element-wise multiplication with a learnable filter \(\mathbf{W} \in \mathbb{C}^{n \times d_t}\):
\begin{equation}
 \tilde{\mathbf{Z}}_i = \mathbf{W} \odot \mathbf{Z}_i, 
\end{equation}
where \(\odot\) denotes the element-wise multiplication. Finally, the MMG utilizes the inverse FFT operation to the modulated spectrum $\tilde{\mathbf{Z}}_i$ back into the time domain:
\begin{equation}
 \tilde{\mathcal{W}}_i = \mathcal{F}^{-1}(\tilde{\mathbf{Z}}_i) \in \mathbb{R}^{n \times d_t}, 
\end{equation}
where \(\mathcal{F}^{-1}(\cdot)\) is the inverse one-dimensional FFT, converting the complex tensor back to a real-valued tensor. To further stabilize training and enhance the embedding, the MMG incorporates a skip connection, layer normalization, and dropout:
\begin{equation}
\hat{\mathcal{W}}_i = \text{LayerNorm}(\tilde{\mathcal{W}}_i + \text{Dropout}(\tilde{\mathcal{W}}_i)).
\end{equation}
Finally, the recovered representation $\hat{\mathcal{W}}_i$ is used as the embedding for the missing modality and is subsequently fed into the pre-trained MMT. 
Additionally, the aforementioned process is applied to scenarios involving missing images to obtain the corresponding vision patch embedding $\hat{\mathcal{V}}_i$.

\begin{table*}[ht]  
\centering  
\resizebox{\textwidth}{!}{
\begin{tabular}{@{}lcccccccccccc@{}}  
\toprule  
& \multicolumn{6}{c}{\textbf{MM-IMDb}} & \multicolumn{3}{c}{\textbf{HateMemes}} & \multicolumn{3}{c}{\textbf{Food101}} \\  
\cmidrule(lr){2-7} \cmidrule(lr){8-10} \cmidrule(lr){11-13}  
\textbf{Missing Type} & \multicolumn{2}{c}{\textbf{Text}} & \multicolumn{2}{c}{\textbf{Image}} & \multicolumn{2}{c}{\textbf{Both}} & \textbf{Text} & \textbf{Image} & \textbf{Both} & \textbf{Text} & \textbf{Image} & \textbf{Both} \\  
\cmidrule(lr){2-3} \cmidrule(lr){4-5} \cmidrule(lr){6-7} \cmidrule(lr){8-8} \cmidrule(lr){9-9} \cmidrule(lr){10-10} \cmidrule(lr){11-11} \cmidrule(lr){12-12} \cmidrule(lr){13-13}  
\textbf{Method} & F1-M & F1-S & F1-M & F1-S & F1-M & F1-S & AUROC & AUROC & AUROC & ACC & ACC & ACC \\  
\midrule   
SMIL & 38.32 & 38.55 & 27.57 & 35.27 & 35.12 & 31.87 & 50.32 & 58.50 & 54.63 & 51.83 & 49.86 & 46.77 \\  
TFR-Net & 37.70 & 38.82 & 38.14 & 39.45 & 37.24 & 38.11 & 51.18 & 55.57 & 52.12 & 65.91 & 67.58 & 63.41 \\  
AcMAE & 47.47 & 46.73 & 43.82 & 42.20 & 44.05 & 43.75 & 55.74 & 59.66 & 57.25 & 69.28 & 73.75 & 71.15 \\ 
\midrule
IF-MMIN & 39.63 & 38.10 & 31.95 & 26.89 & 31.98 & 29.33 & 57.62 & 53.44 & 55.19 & 66.76 & 64.36 & 68.53 \\  
ShaSpec & 44.04 & 42.05 & 44.23 & 42.53 & 44.06 & 42.13 & 58.75 & 60.30 & \underline{60.96} & 60.99 & 74.87 & 70.02 \\  
DrFuse & 47.05 & 45.22 & 43.58 & 42.19 & \underline{48.83} & \underline{47.15} & 57.60 & \underline{60.66} & 55.84 & 66.30 & 75.09 & 68.23 \\  
CorrKD & 44.82 & 45.27 & 39.48 & 39.11 & 41.20 & 40.51 & 58.74 & 55.59 & 57.91 & 61.37 & 66.83 & 62.87 \\  
\midrule
MAPs & 46.12 & 45.47 & \underline{44.86} & \underline{43.19} & 45.48 & 44.30 & 58.62 & 60.16 & 58.89 & 67.02 & 75.62 & 72.52 \\  
MSPs & \underline{49.16} & \underline{48.81} & 44.62 & 43.06 & 48.28 & 46.71 & \underline{59.60} & 60.05 & 59.08 & \underline{71.74} & \underline{79.09} & \underline{74.46} \\
\midrule
\textbf{\M} & \textbf{55.16} & \textbf{55.00} & \textbf{46.44} & \textbf{45.12} & \textbf{50.89} & \textbf{50.22} & \textbf{64.10} & \textbf{62.57} & \textbf{63.47} & \textbf{75.53} & \textbf{81.98} & \textbf{76.94} \\  
Improv. (\%) & 12.21$\uparrow$ & 12.68$\uparrow$ & 3.52$\uparrow$ & 4.47$\uparrow$ & 4.22$\uparrow$ & 6.51$\uparrow$ & 7.55$\uparrow$ & 3.15$\uparrow$ & 4.12$\uparrow$ & 5.28$\uparrow$ & 3.65$\uparrow$ & 3.33$\uparrow$ \\  
\textit{p}-val. & $8.93e^{-6}$ & $1.73e^{-5}$ & $5.94e^{-5}$ & $9.68e^{-6}$ & $6.43e^{-6}$ & $2.92e^{-5}$ & $1.24e^{-6}$ & $3.44e^{-5}$ & $1.03e^{-5}$ & $1.63e^{-6}$ & $3.24e^{-6}$ & $8.50e^{-5}$ \\  
\bottomrule  
\end{tabular}  
}  
\caption{Performance comparison on three datasets with a 70\% missing rate across various missing-modality scenarios. The best results are in \textbf{bold} font and the second \underline{underlined}. Higher values of F1-M, F1-S, AUROC, and ACC indicate better performance.}  
\label{tab:performance}   
\end{table*}

\begin{table}[t]
  \centering
  \resizebox{\linewidth}{!}{
  \begin{tabular}{lrrrrr}
    \toprule
    \textbf{Dataset}&\textbf{\# Image}&\textbf{\# Text}&\textbf{\# Train}&\textbf{\# Val}&\textbf{\# Test}\\
    % \hline
    \midrule
    MM-IMDb	&25,959  &25,959  &15,552 &2,608 &7,799 \\
    HateMemes &10,000 &10,000 &8,500 &500 &1,500 \\
    Food101 & 90,688 & 90,688 & 67,972 & - & 22,716  \\
    \bottomrule
  \end{tabular}
  }
     \caption{Statistics of three multimodal downstream datasets.}
  \label{tab:dataset}
\end{table}

\noindent \textbf{Dynamic Prompt-Tuning}
Given a pre-trained MMT $f_{\theta}$ with $N$ consecutive Multi-head Self-Attention (MSA) layers, we denote the input representation of the $b$-th MSA layer as  $\mathbf{h}^b \in \mathbb{R}^{L \times d},\ b = 1, 2, \ldots, N$ with input length $L$ and embedding dimension $d$. For full-modality data, we utilize the embedding layer of the pre-trained model $f_{\theta}(\cdot)$ to obtain the corresponding text embedding $\mathbf{E}^t$ and image embedding $\mathbf{E}^v$. In the case of missing-modality, we employ the generated word embedding $\hat{\mathcal{W}}$ and vision patch embedding $\hat{\mathcal{V}}$ to fill the corresponding missing modality. $\mathbf{h}^1$ is the concatenation of text embedding $\mathbf{E}^t$ and image embedding $\mathbf{E}^v$. 
The context-aware prompts $\mathbf{P}^{t}$, $\mathbf{P}^{v}$, and $\mathbf{P}^{l}$ are then attached to the embedding features along the sequence-length dimension to form the extended features $\mathbf{h}_p^b = [\mathbf{P}^{t}, \mathbf{P}^{v}, \mathbf{P}^{l}, \mathbf{h}^b]$. These extended features $\mathbf{h}_p^{b}$ are fed into the MMT starting from the $b$-th layer and continue to propagate through the remaining layers. The final output $\mathbf{h}_{p}^{N}$ represents comprehensive modal representation after the $N$-th layer.
Rather than adding prompts at each MSA layer, which can result in considerable overhead, we selectively insert the prompts into the specific $b$-th layer. 

\noindent \textbf{Label-Augmented Prediction} To further leverage the contextual information in label-level prompts, we design a label-augmented classifier by computing the similarity between the output representation of the MMT and the label matrix $\tilde{\mathbf{P}}^{l}$. Specifically, for the final prediction, we feed the output representation $\mathbf{h}_{p}^{N}$ into the pooler layer to obtain the representation $\bm{Z} \in \mathbb{R}^{d \times 1}$. Next, we calculate the probabilities $\hat{\mathbf{y}}\in\mathbb{R}^{C \times 1}$ for $C$ classes: $\hat{\mathbf{y}} = \text{softmax} (\tilde{\mathbf{P}}^{{l}}{*}\bm{Z})$. During training, we freeze all parameters in the MMT and optimize the model using cross-entropy loss.

%% file: 5-Experiment.tex
\subsection{Experimental Settings}
A summary of the experimental settings is provided in this section, which refers to datasets, baselines, evaluation metrics, setting of missing pattern, and implementation details.

\noindent \textbf{Datasets} 
Following previous work~\cite{lee2023multimodal, jang2024towards}, we evaluate our \M~on three downstream tasks. (1) MM-IMDb \cite{arevalo2017gated}, primarily used for movie genre classification involving both image and text modalities. (2) Food101 \cite{wang2015recipe}, which focuses on image classification that incorporates both image and text. (3) HateMemes \cite{kiela2020hateful}, aimed to identify hate speech in memes using image and text modalities. 
Detailed statistics of datasets are presented in Table \ref{tab:dataset}. 
The dataset splits are consistent with the original paper.

\noindent \textbf{Baselines}
We compare our \M~with 9 competitive baselines, which are classified into three categories: (1) \textit{Cross-modal generation methods}: SMIL \cite{ma2021smil}, TFRNet \cite{yuan2021transformer}, and AcMAE \cite{woo2023towards}. (2) \textit{Joint learning methods}: IF-MMIN \cite{zuo2023exploiting}, ShaSpec \cite{wang2023multi}, DrFuse \cite{yao2024drfuse}, and CorrKD~\cite{li2024correlation}.  (3) \textit{Prompt-based methods}: MAPs \cite{lee2023multimodal} and MSPs \cite{jang2024towards}. 

\noindent \textbf{Evaluation Metrics} 
Following prior works \cite{lee2023multimodal, jang2024towards}, we adopt appropriate dataset-specific metrics for evaluation: F1-Micro (F1-M) and F1-Sample (F1-S) for the MM-IMDb dataset, AUROC for the HateMemes dataset, and classification accuracy (ACC) for the Food101 dataset.

\noindent \textbf{Setting of Missing Pattern} 
We assume training set is fully available and define the missing rate $\eta\%$ as the rate of modality-incomplete samples in the test set: (1) text/image missing with $\eta\%$ indicates that there are $\eta\%$ image-only/text-only instances and (1-$\eta\%$) modality-complete instances. (2) both modalities missing with $\eta\%$ indicates that there are $\frac{\eta}{2}\%$ text-only instances, $\frac{\eta}{2}\%$ image-only instances and (1-$\eta\%$) modality-complete instances. 
We set missing rate $\eta\%$ = 70\% by default.
For training of each model, we simulate the same 70\% missing rate to align model optimization well with test conditions, but allow each model to access the full modality information in the training set.

% We define the missing rate $\eta\%$ as the proportion of modality-incomplete data relative to the entire dataset. 
% For each dataset, there are three possible cases of missing-modality: text missing, image missing, and both modalities missing. Text/image missing with a missing rate of $\eta\%$ indicates that there are $\eta\%$ instances consisting of texts/images and (1-$\eta$)\% instances that contain both modalities. Missing both modalities with a missing rate of $\eta\%$ indicates that there are $\frac{\eta}{2}\%$ instances consisting solely of images, $\frac{\eta}{2}\%$ instances consisting solely of text, and (1-$\eta$)\% instances that are complete, containing both modalities.

\noindent \textbf{Implementation Details} 
Following prior works \cite{lee2023multimodal, jang2024towards}, we utilize the pre-trained ViLT \cite{kim2021vilt} as our MMT backbone. The memory $\mathcal{B}$ for each dataset is constructed with the corresponding training and validation set without test data leakage. The length $l$ of prompts is set to 2, the number of retrieved instances $K$ is chosen from $\{1, 3, 5, 7, 9\}$, and the prompt insertion layer $b$ is set to 2. We utilize the AdamW optimizer with a learning rate of $1 \times 10^{-3}$ and total 20 epochs for optimizing the parameters. All experiments are conducted with an NVIDIA RTX 3090 GPU. 

\subsection{Overall Performance}
To verify the superiority of \M, we compare it with 9 competitive baselines on three datasets under a missing rate of $\eta\% = 70\%$. We have the following observations:

First, our \M~consistently outperforms all strong baselines on three datasets under various modal conditions and metrics. 
Moreover, we retrain \M~and the best-performing baseline five times to calculate the $p$-value. 
Notably, \M~achieves improvements of 12.21\% and 12.68\% in the F1-M and F1-S metrics, respectively, on the MM-IMDb dataset with missing text. 
These results validate our design of exploiting expressive knowledge from retrieved instances to enhance both missing and complete modality data.  
Meanwhile, the missing modality generator and context-aware prompter distill expressive contextual information from retrieved instances to approximate missing content and generate dynamic prompts, respectively, thereby improving model robustness for incomplete modalities. 

Second, \textit{cross-modal generation} and \textit{joint learning methods} demonstrate inferior performance, primarily due to the uncertainty introduced by random placeholders and the challenges of modality heterogeneity in reconstruction, which create significant performance bottlenecks. 
Moreover, \textit{prompt-based methods} also exhibit limited effectiveness in missing-modality scenarios, as they rely on dummy imputations and static prompting strategies, further restricting their potential and resulting in performance stagnation.

\begin{table}[t]
    \centering
    \setlength{\tabcolsep}{4pt}
    \resizebox{\linewidth}{!}{
    \begin{tabular}{cccccc}
        \toprule
        \multirow{2}{*}{\textbf{Module}} & \multirow{2}{*}{\textbf{Variant}} & \multicolumn{2}{c}{\textbf{MM-IMDb}} & {\textbf{HateMemes}} & {\textbf{Food101}} \\ 
        & & \textbf{F1-M} & \textbf{F1-S} & \textbf{AUROC} & \textbf{ACC} \\
         \midrule
        \textbf{\M} & \textbf{All} & \textbf{55.16} & \textbf{55.00} & \textbf{64.10} & \textbf{75.53} \\
        \midrule
        \multirow{2}{*}{Retriever}
        & CM Retriever & 52.37 & 51.70 & 61.87 & 74.24 \\
        & w/o Retriever & 49.25 & 48.36 & 60.29 & 73.60 \\   
        \midrule
        \multirow{2}{*}{Generator}
        & Padding & 51.14 & 51.63 & 61.30 & 72.78 \\
        & w/o Filter & 54.15 & 52.99 & 60.67 & 74.07  \\
        \midrule
        \multirow{3}{*}{Prompter}
        & Static Prompt & 54.38 & 53.14 & 62.65 & 74.40 \\
        & w/o Label & 53.41 & 53.45 & 62.01 & 74.34  \\
        & w/o Prompter & 51.49 & 50.43 & 60.94 & 72.65 \\
        \bottomrule
    \end{tabular}
    }
    \caption{Ablation study of \M~under 70\% text missing.}
    \label{table:ablation1}
\end{table}

\subsection{Ablation Study}
We conduct various ablation experiments to evaluate the impact of each component within \M~under a 70\% text missing case and summarize the results in Table \ref{table:ablation1}. 

\noindent \textbf{Effect of Multi-Channel Retriever} To analyze the impact of the retriever in \M, we designed two variants: (1) \textbf{CM Retriever}: replacing the multi-channel retriever with cross-modal retriever, 
and (2) \textbf{w/o Retriever}: removing the retriever entirely. These results confirm the presence of the modal gap problem in cross-modal retrieval, which renders the retrieved instances irrelevant to the target images. Furthermore, this finding reinforces our design of the multi-channel retrieval that retrieves relevant instances by calculating within-modality similarities, thereby enhancing both missing and complete modality data.

\noindent \textbf{Effect of Missing Modality Generator} To evaluate the impact of the missing modality generator, we designed variant models: (1) \textbf{Padding}: using random values to fill in the missing modality, and (2) \textbf{w/o Filter}: removing the filter block entirely. We observe that dummy padding results in a decline in performance.
This finding supports our assertion that dummy padding contributes to performance bottlenecks in prompt-based methods. Additionally, the removal of the filter layer leads to a significant performance drop, underscoring the importance of the filter layer in \M~for effectively mitigating noise. 

\noindent \textbf{Effect of Context-Aware Prompter} To analyze the context-aware prompts,
we design variants: (1) \textbf{Static Prompt}: replacing context-aware prompts with static prompts; (2) \textbf{w/o Label}: removing label enhancement; and (3) \textbf{w/o Prompter}: eliminating text-, vision-, label-prompts entirely. The three variants result in poorer performance, validating that static prompts offer limited relevant cues for addressing incomplete multimodal learning. 

\begin{figure}[t]
    \centering     
    \subfloat[Effect of $K$ on MM-IMDb]{\includegraphics[width=0.495\columnwidth]{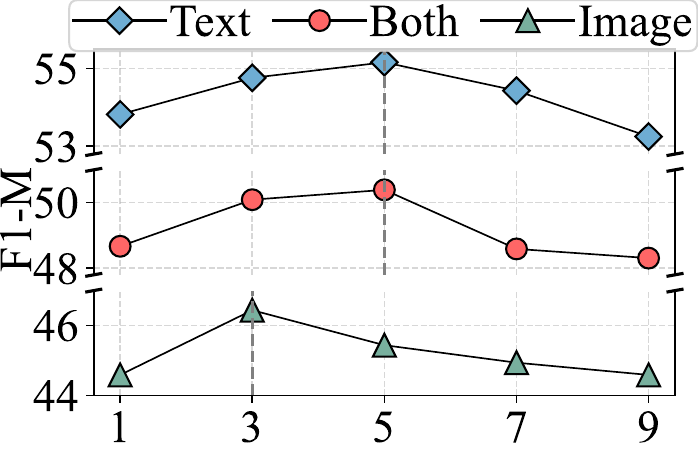}} 
    \subfloat[Effect of $K$ on HateMemes]{\includegraphics[width=0.495\columnwidth]{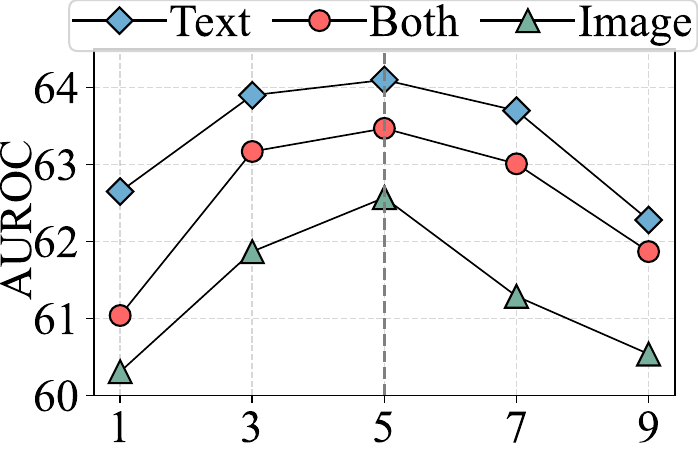}} 
      \caption{Hyper-parameter analysis of $K$ under three modality missing scenarios.}
      \label{fig:sensitive}
\end{figure}

\subsection{Hyper-Parameter Analysis} 
Fig.~\ref{fig:sensitive}(a) and \ref{fig:sensitive}(b) present the sensitivity analysis of \M’s hyper-parameters $K$ on the MM-IMDb and HateMemes datasets. The results demonstrate that the performance of \M~is improved by retrieving relevant instances. However, incorporating a larger number of instances may result in a decline in performance due to the introduction of noise (i.e., irrelevant instances). 
Consequently, we adopt $K=3$ under the image missing case on the MM-IMDb dataset and $K=5$ under other scenarios.

\begin{figure}[t]
  \centering
  \includegraphics[width=1\linewidth]{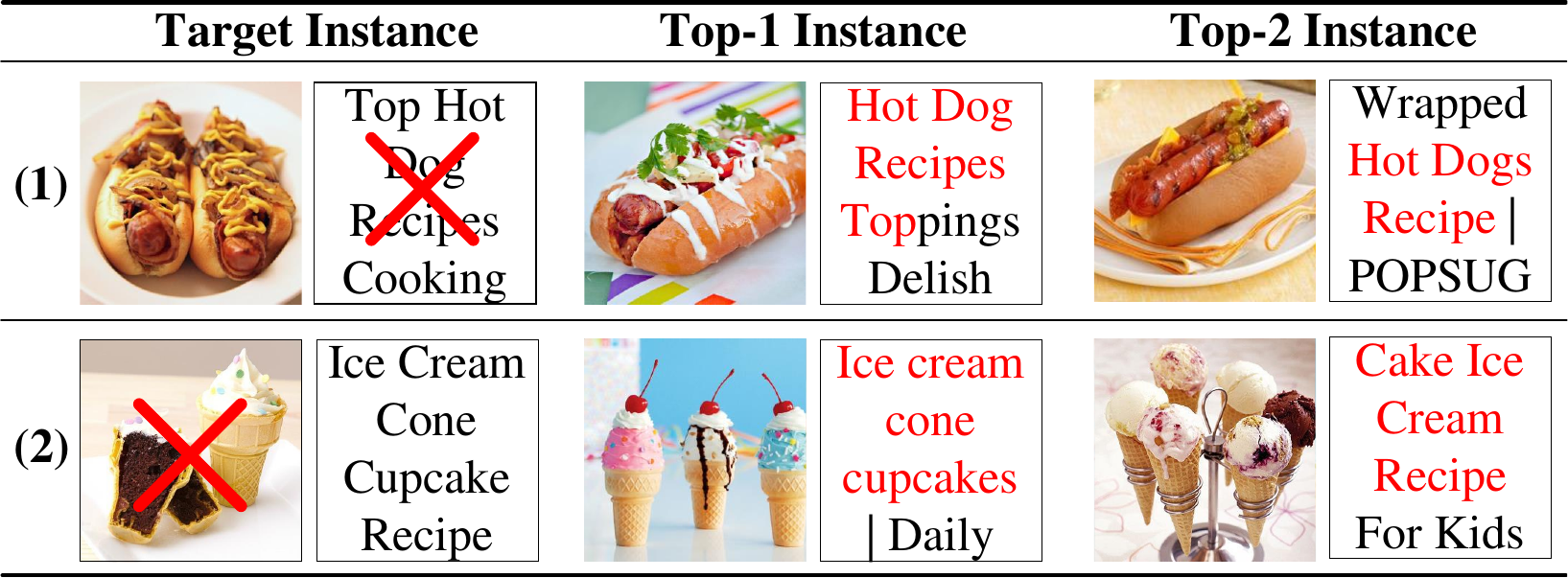}
\caption{Examples of Top-2 retrieved instances for two modality-incomplete target instances. The first target instance is image-only while the second one is text-only.  \rc{Red} texts highlight similar content.}
\label{fig:case-retrieval}
\end{figure}

\subsection{Retrieval Quality Presentation} 
To further analyze the efficacy of our proposed multi-channel retriever, we randomly select two instances with incomplete modalities from the Food101 dataset. Fig.~\ref{fig:case-retrieval} visualizes the Top-2 similar retrieved instances, demonstrating a strong semantic correlation between the retrieved and target instances in both image and text modalities. The high quality of retrieval relevance indicates our multi-channel retriever's ability to effectively identify relevant modal information.

\begin{figure}[t]
    \centering
    \subfloat[Text Missing]{\includegraphics[width=0.495\columnwidth]{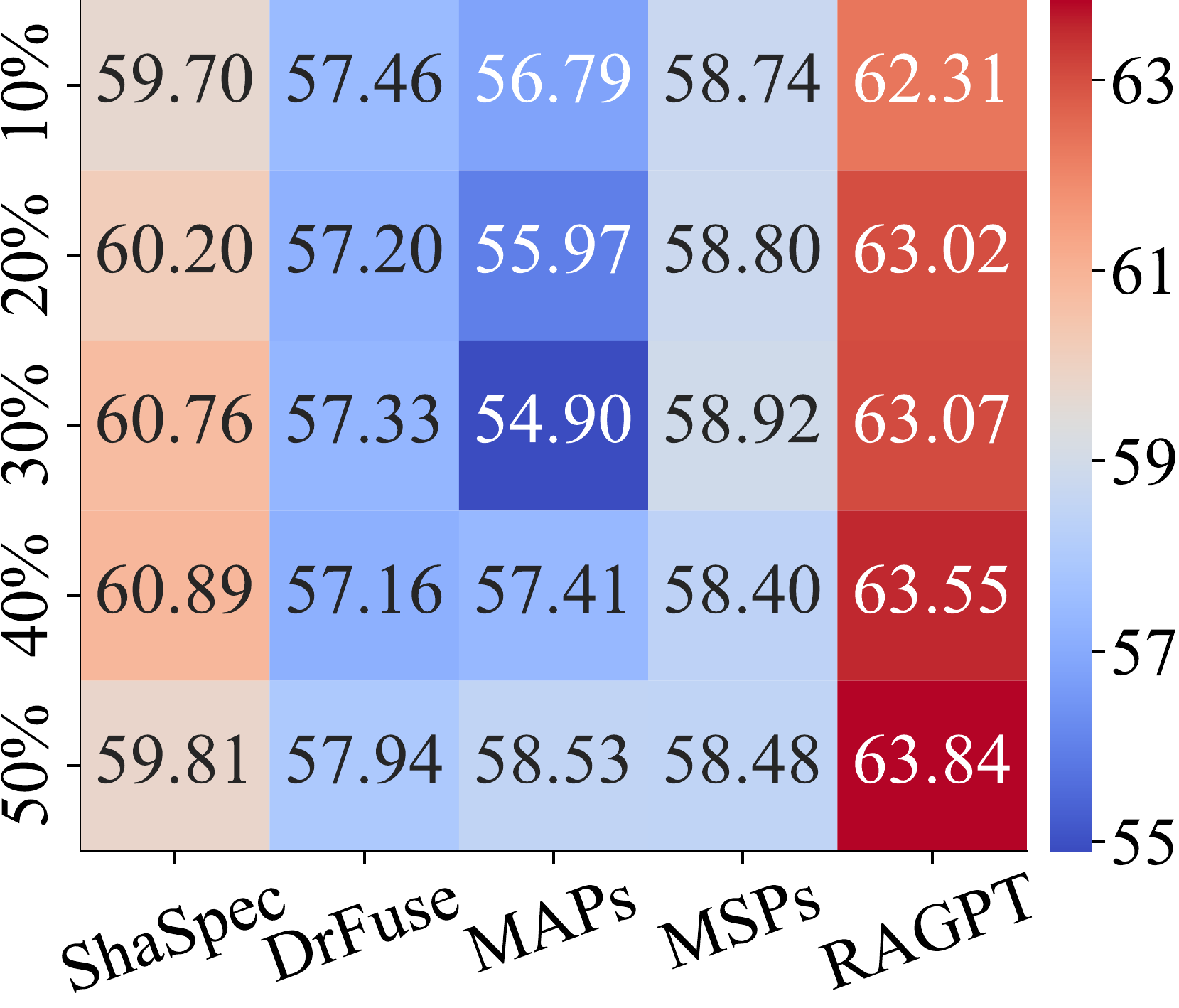}} \hspace{0.4mm}
    \subfloat[Both Missing]{\includegraphics[width=0.484\columnwidth]{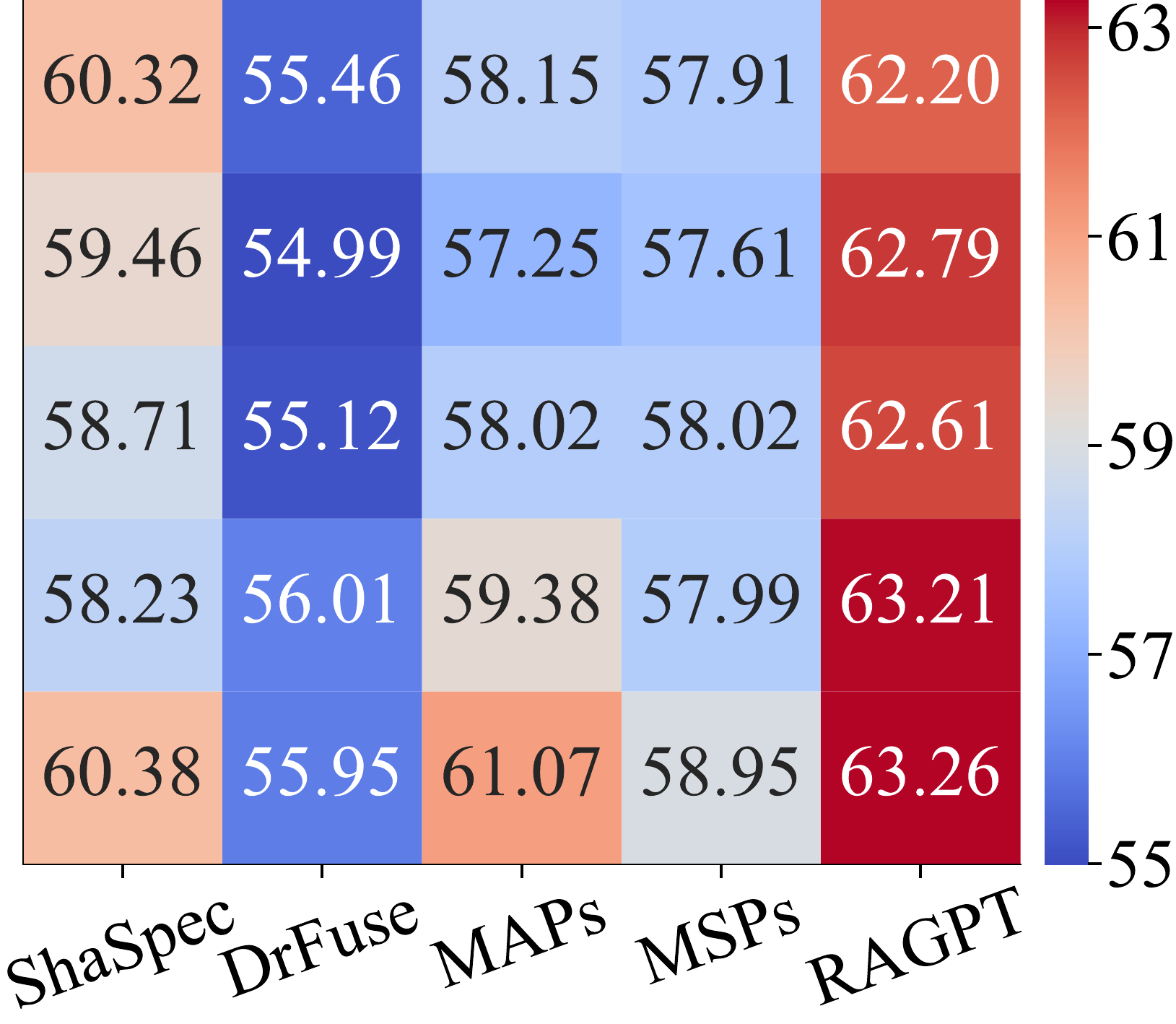}}
    \caption{Generalization analysis on the HateMemes dataset across various missing rates in terms of AUROC.}
    \label{fig:gneralizability}
\end{figure}

\subsection{Model Generalizability}
To investigate the model's generalizability, we design two experiments with varying missing rates in the training set and evaluate their performance on a test set with a 90\% missing rate. Compared with four strong baselines (ShaSpec, DrFuse, MAPs, and MSPs), Fig.~\ref{fig:gneralizability}(a) shows the results for the missing-text case, while Fig.~\ref{fig:gneralizability}(b) presents the results for scenarios of missing both modalities. We observe that our \M~outperforms all baselines across all missing rates, demonstrating superior performance for missing-modality. These results highlight \M's generalizability, which can be attributed to the ability of exploring crucial cues from relevant contexts. 

\begin{figure}[t]
    \centering
  \includegraphics[width=1\linewidth]{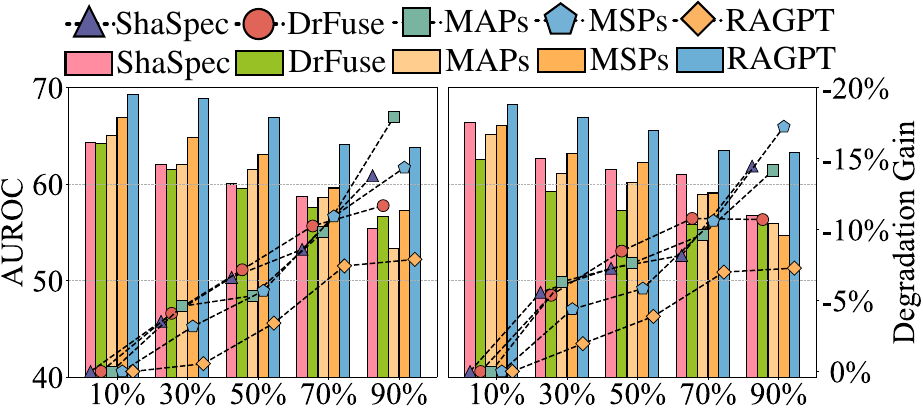}
    \caption{Robustness analysis on the HateMemes dataset across various missing rates in terms of AUROC.}
    \label{fig:robustness}
\end{figure}

\subsection{Robustness to Different Missing Rates} 
We conduct an experiment to analyze the model's robustness to varying missing rates. Fig.~\ref{fig:robustness} illustrates the results comparing \M~with four strong baselines (ShaSpec, DrFuse, MAPs, and MSPs) on the HateMemes dataset. We observe that 
the performance of all baselines deteriorates markedly as the missing rate increases. 
In contrast, \M~demonstrates only a slight performance decrease as the missing rate increases. This result highlights the valuable components of \M~for effectively mitigating the impact of missing data. 
Specifically, \M~leverages expressive knowledge from retrieved instances to approximate missing modalities through the missing modality generator. Additionally, \M~generates context-aware prompts that enhance the performance of the pre-trained MMTs.

\begin{figure}[t]
    \centering
    \subfloat[Text Missing on MAPs]{\includegraphics[width=0.512\columnwidth]{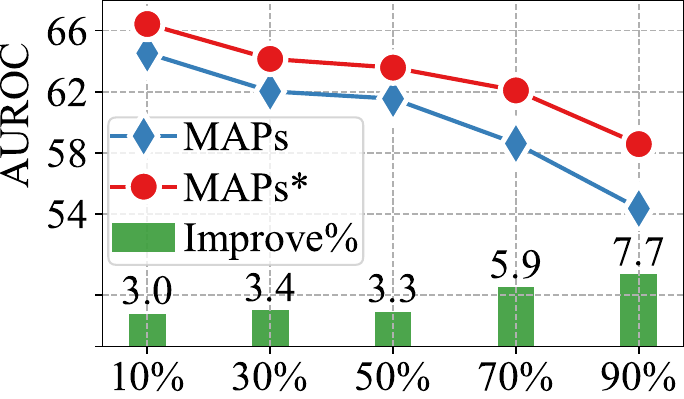}} \hspace{0.1mm}
     \subfloat[Text Missing on MSPs]{\includegraphics[width=0.477\columnwidth]{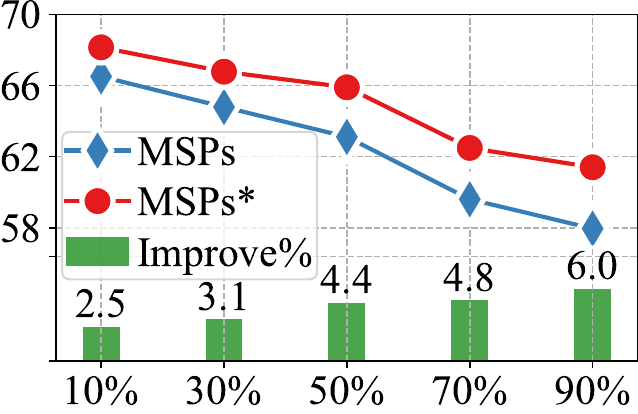}}
    \caption{Effect of integrating key modules in \M~for baselines on the HateMemes dataset in terms of AUROC.}
    \label{fig:insertion}
\end{figure}

\subsection{Model Scalability}
To further validate the \M's scalability, we integrate key modules (multi-channel retriever, missing modality generator, and context-aware prompter) into two prompt-based baselines (MAPs and MSPs). In Fig.~\ref{fig:insertion}, we observe a significantly slower rate of performance decline in the two baselines as the missing rate increased. This finding indicates that our modules significantly enhance the robustness of these baselines for incomplete modalities. It also validates the effectiveness of our design in extracting informative multimodal cues from relevant instances and prompting pre-trained MMTs.

\begin{figure}[t]
    \centering
    \subfloat[\M]{\includegraphics[width=0.49\columnwidth]{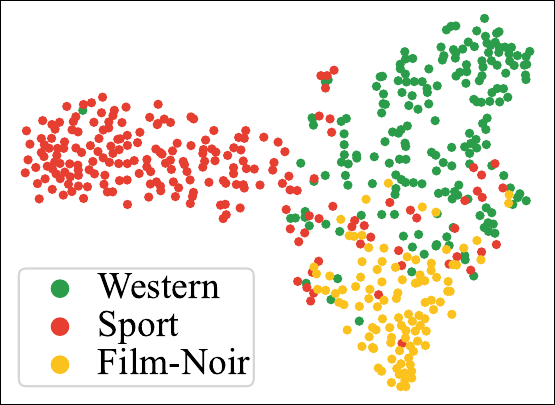}}\hspace{0.25mm}
    \subfloat[MSPs]{\includegraphics[width=0.49\columnwidth]{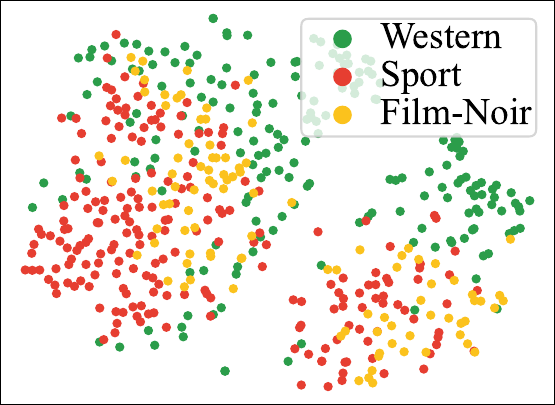}}
    \caption{t-SNE visualization of \M~and MSPs on the MM-IMDb dataset under a 90\% text missing rate.}
    \label{fig:tsne}
\end{figure}

\subsection{Model Prediction Visualization}
Fig.~\ref{fig:tsne} illustrates the t-SNE \cite{van2008visualizing} visualization of the embedding distributions for three genres (i.e., Sport, Film-Noir, and Western) in the MM-IMDb test set under a 90\% text missing rate.
We observe that while baseline MSPs learns distinguishable features, the learned features remain intertwined. In contrast, the representations of three genres learned by our \M~are more discriminative, exhibiting larger segregated areas among instances with different labels.

%% file: 6-Conclusion.tex
In this work, we proposed \M, a novel retrieval-augmented dynamic prompt-tuning framework to address the missing-modality issue. This model-agnostic framework includes three key components: (1) the multi-channel retriever, (2) the missing modality generator, and (3) the context-aware prompter, to effectively inject valuable contextual knowledge into pre-trained MMT, thereby enhancing its robustness in the missing-modality scenario. Extensive experiments conducted on three real-world datasets demonstrate the superiority of \M~in tackling incomplete modality learning.